\documentclass[letterpaper, 10 pt, conference]{ieeeconf}  % Comment this line out if you need a4paper

\IEEEoverridecommandlockouts                              % This command is only needed if 
\usepackage{color,xcolor}
\usepackage{epsfig}
\usepackage{graphicx}

\usepackage{adjustbox}
\usepackage{array}
\usepackage{booktabs}
\usepackage{colortbl}
\usepackage{float,wrapfig}
\usepackage{hhline}
\usepackage{multirow}
\usepackage{subcaption} % issues a warning with CVPR/ICCV format
\usepackage[font={small}]{caption}

\usepackage{amsmath,amsfonts,amssymb}
\usepackage{bm}
\usepackage{nicefrac}
\usepackage{microtype}

\usepackage{changepage}
\usepackage{extramarks}
\usepackage{fancyhdr}
\usepackage{lastpage}
\usepackage{setspace}
\usepackage{soul}
\usepackage{xspace}

\usepackage{url}

\usepackage{algorithm, algorithmic}
\usepackage{enumerate}
\usepackage{todonotes} % conflict with CVPR/ICCV format
\usepackage{gensymb}
\usepackage{breqn}
\newcolumntype{L}[1]{>{\raggedright\let\newline\\\arraybackslash\hspace{0pt}}m{#1}}
\newcolumntype{C}[1]{>{\centering\let\newline\\\arraybackslash\hspace{0pt}}m{#1}}
\newcolumntype{R}[1]{>{\raggedleft\let\newline\\\arraybackslash\hspace{0pt}}m{#1}}

\newcommand{\sect}[1]{Sec.~\ref{#1}}

\newcommand{\fig}[1]{Fig.~\ref{#1}}

\newcommand{\tbl}[1]{Table~\ref{#1}}

\newcommand{\ignorethis}[1]{}

\DeclareMathOperator*{\argmin}{arg\,min}

\makeatletter
\DeclareRobustCommand\onedot{\futurelet\@let@token\@onedot}
\def\@onedot{\ifx\@let@token.\else.\null\fi\xspace}

\def\eg{e.g\onedot} 
\def\ie{i.e\onedot}

\def\etal{et al\onedot}
\makeatother

\definecolor{MyDarkBlue}{rgb}{0,0.08,1}
\definecolor{MyDarkGreen}{rgb}{0.02,0.6,0.02}
\definecolor{MyDarkRed}{rgb}{0.8,0.02,0.02}
\definecolor{MyDarkOrange}{rgb}{0.40,0.2,0.02}
\definecolor{MyPurple}{RGB}{111,0,255}
\definecolor{MyRed}{rgb}{1.0,0.0,0.0}
\definecolor{MyGold}{rgb}{0.75,0.6,0.12}
\definecolor{MyDarkgray}{rgb}{0.66, 0.66, 0.66}

\newcommand{\Modelfull}{Simulator-Augmented Interaction Networks\xspace}
\newcommand{\modelfull}{simulator-augmented interaction networks\xspace}
\newcommand{\model}{SAIN\xspace}

\title{\LARGE \bf
Combining Physical Simulators and Object-Based Networks for Control
}

\author{Anurag Ajay$^{1}$, Maria Bauza$^{2}$, Jiajun Wu$^{1}$, Nima Fazeli$^{2}$,\\Joshua B. Tenenbaum$^{1}$, Alberto Rodriguez$^{2}$, Leslie P. Kaelbling$^{1}$% <-this % stops a space
\thanks{$^{1}$Anurag Ajay, Jiajun Wu, Joshua B. Tenenbaum, and Leslie P. Kaelbling are with the Computer Science and Artificial Intelligence Laboratory (CSAIL) at Massachusetts Institute of Technology, Cambridge, MA, USA}%
\thanks{$^{2}$Maria Bauza, Nima Fazeli, and Alberto Rodriguez are with the Department of Mechanical Engineering at Massachusetts Institute of Technology, Cambridge, MA, USA}%
}

\begin{document}

\maketitle
\thispagestyle{empty}
\pagestyle{empty}

%%%%%%%%%%%%%%%%%%%%%%%%%%%%%%%%%%%%%%%%%%%%%%%%%%%%%%%%%%%%%%%%%%%%%%%%%%%%%%%%
\begin{abstract}
Physics engines play an important role in robot planning and control; however, many real-world control problems involve complex contact dynamics that cannot be characterized analytically. Most physics engines therefore employ approximations that lead to a loss in precision. In this paper, we propose a hybrid dynamics model, \modelfull (\model), combining a physics engine with an object-based neural network for dynamics modeling. %Since learning residual between the physics engine and real world is a simpler problem, our models are more data-efficient. 
Compared with existing models that are purely analytical or purely data-driven, our hybrid model captures the dynamics of interacting objects in a more accurate and data-efficient manner. Experiments both in simulation and on a real robot suggest that it also leads to better performance when used in complex control tasks. Finally, we show that our model generalizes to novel environments with varying object shapes and materials.
\end{abstract}
\section{Introduction}
\label{sec:intro}

Physics engines are important for planning and control in robotics. To plan for a task, a robot may use a physics engine to simulate the effects of different actions on the environment and then select a sequence of them to reach a desired goal state. The utility of the resulting action sequence depends on the accuracy of the physics engine's predictions, so a high-fidelity physics engine is an important component in robot planning. Most physics engines used in robotics (such as Mujoco~\cite{Todorov2012MuJoCo} and Bullet~\cite{Coumans:2015}) use approximate contact models, and recent studies \cite{Kolbert2016,Yu2016More,Fazeli2017Fundamental} have demonstrated discrepancies between their predictions and real-world data. These mismatches make contact-rich tasks hard to solve using these physics engines.

One way to increase the robustness of controllers and policies resulting from physics engines is to add perturbations to parameters that are difficult to estimate accurately (\eg, frictional variation as a function of position \cite{Yu2016More}). This approach leads to an ensemble of simulated predictions that covers a range of possible outcomes. Using the ensemble allows to take more conservative actions and increases robustness, but does not address the limitation of using learned, approximate models~\cite{mordatch2015ensemble,becker2012approximate}.

To correct for model errors due to approximations, we learn a residual model between real-world measurements and a physics engine's predictions. Combining the physics engine and residual model yields a {\em data-augmented physics engine}. This strategy is effective because learning a residual error of a reasonable approximation (here from a physics engine) is easier and more sample efficient than learning from scratch. This approach has been shown to be more data efficient, have better generalization capabilities, and outperform its purely analytical or data-driven counterparts~\cite{Fazeli2017Learning,ajay2018augmenting, chatzilygeroudis2017using, kloss2017combining}. 

%\nf{I think the first few lines of the paragraph are a little hard to understand. Consider: One drawback to prior approaches is that they are limited to fixed-length vectors of parameter values (still not sure what you mean by a fixed-length vector of parameter values -- parameters for what?). As a consequence, they cannot handle changes in the number of objects and do not generalize what they learn for a single object to multiple other similar objects. }
Most residual-based approaches assume a fixed number of objects in the world states. This means they cannot be applied to states with a varied number of objects or generalize what they learn for one object to other similar ones. This problem has been addressed by approaches that use graph-structured network models, such as interaction networks~\cite{Battaglia2016Interaction} and neural physics engines~\cite{Chang2017compositional}. These methods are effective at generalizing over objects, modeling interactions, and handling variable numbers of objects. However, as they are purely data-driven, in practice they require a large number of training examples to arrive at a good model.  

\begin{figure}[t]
    \centering
    \includegraphics[width=\linewidth]{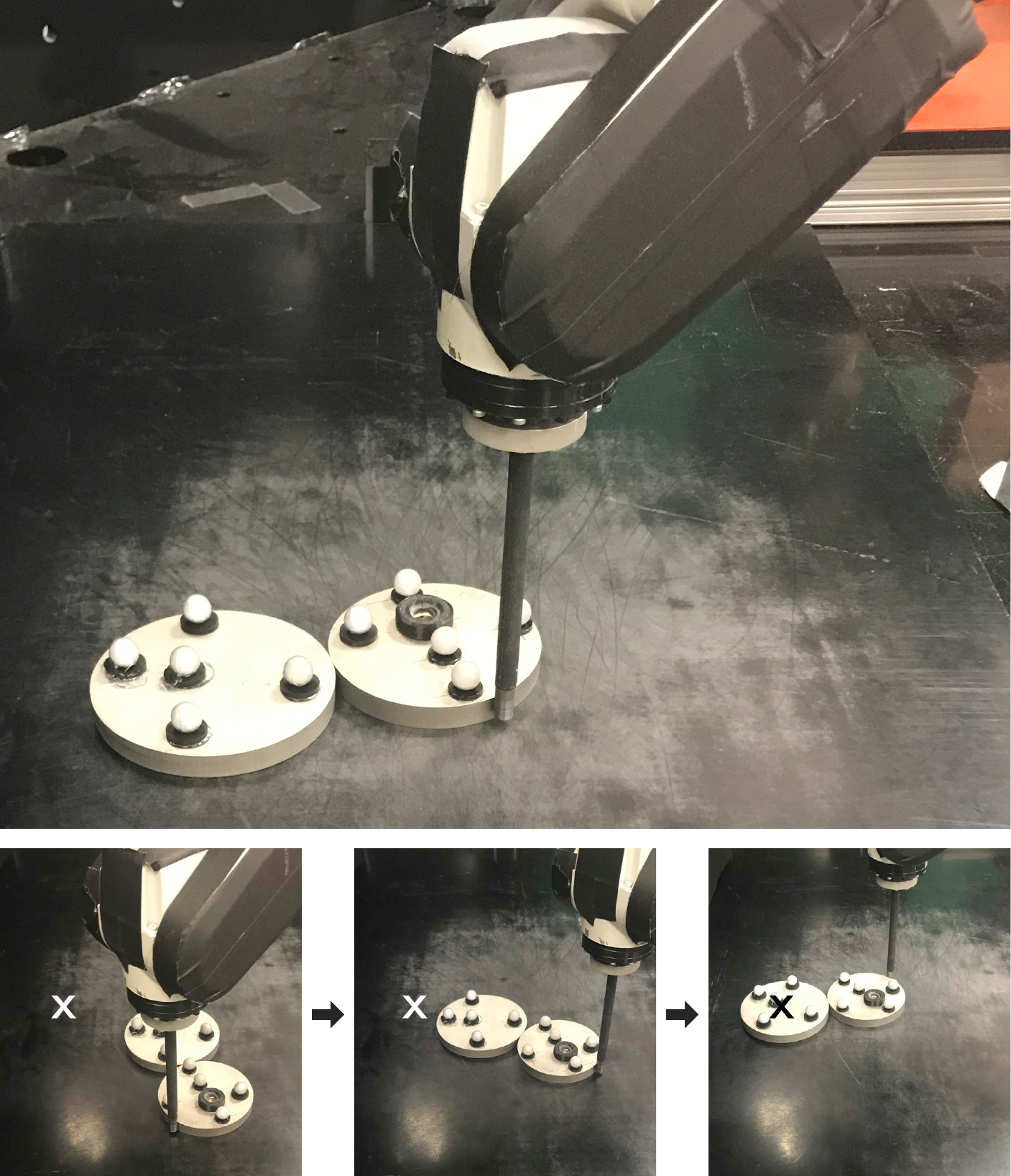}
    \caption{\textbf{Top}: the robot wants to push the second disk to a goal position by pushing on the first disk. \textbf{Bottom}: three snapshots within a successful push (target marked as X). The robot learns to first push the first disk to the right and then use it to push the second disk to the target position.}
    \label{fig:teaser}
    \vspace{-15pt}
\end{figure}

% \begin{figure}[t]
%     \centering
%     \includegraphics[width=\linewidth]{fig/short_teaser.pdf}
%     \caption{Three snapshots within a successful push (target marked as X). The robot learns to first push the first disk to the right and then use it to push the second disk to the target position.}
%     \label{fig:teaser}
%     \vspace{-15pt}
% \end{figure}

In this paper, we propose \modelfull (\model), incorporating interaction networks into a physical simulator for complex, real-world control problems. Specifically, we show:
\begin{itemize}
    \item Sample-efficient residual learning and improved prediction accuracy relative to the physics engine,
    \item Accurate predictions for the dynamics and interaction of novel arrangements and numbers of objects, and the
    \item Utility of the learned residual model for control in highly underactuated planar pushing tasks.
\end{itemize}

We demonstrate \model's performance on the experimental setup depicted in \fig{fig:teaser}. Here, the robot's objective is to guide the second disk to a goal by pushing on the first. This task is challenging due to the presence of multiple complex frictional interactions and underactuation~\cite{hogan2016feedback}. We demonstrate the step-by-step deployment of \model, from training in simulation to augmentation with real-world data, and finally control.

%To this end, we first train an interaction networks on an ensemble of simulations. Using this learned model, we perform the task and record the pushing data. Now, we use this real-world data to fine-tune our model as well as to learn an augmented interaction-network model. We show that \model outperforms the other models in terms of prediction error as well as generating accurate control sequences for solving the task.

\section{Related Work}
\label{sec:related}

\subsection{Learning Contact Dynamics}

In the field of contact dynamics, researchers have looked towards data-driven techniques to complement analytical models and/or directly learn dynamics. For example, Byravan and Fox~\cite{byravan2017se3} designed neural nets to predict rigid-body motions for planar pushing. Their approach does not exploit explicit physical knowledge. Kloss~\etal~\cite{kloss2017combining} used neural net predictions as input to an analytical model; the output of the analytical model is used as the prediction. Here, the neural network learns to maximize the analytical model's performance. Fazeli~\etal~\cite{Fazeli2017Learning} also studied learning a residual model for predicting planar impacts. Zhou~\etal~\cite{zhou2016convex} employed a data-efficient algorithm to capture the frictional interaction between an object and a support surface. They later extended it for simulating parametric variability in planar pushing and grasping~\cite{zhou2017fast}.

The paper closest to ours is that from Ajay~\etal~\cite{ajay2018augmenting}, where they used the analytical model as an approximation to the push outcomes, and learned a residual neural model that makes corrections to its output. In contrast, our paper makes two key innovations: first, instead of using a feedforward network to model the dynamics of a single object, we employ an object-based network to learn residuals. Object-based networks build upon explicit object representations and learn how they interact; this enables capturing multi-object interactions. Second, we demonstrate that such a hybrid dynamics model can be used for control tasks both in simulation and on a real robot.

\subsection{Differentiable Physical Simulators}

There has been an increasing interest in building differentiable physics simulators~\cite{Ehrhardt2017Taking}. For example, Degrave~\etal~\cite{degrave2016differentiable} proposed to directly solve differentiable equations. Such systems have been deployed for manipulation and planning for tool use~\cite{toussaint2018differentiable}. Battaglia~\etal~\cite{Battaglia2016Interaction} and Chang~\etal~\cite{Chang2017compositional} have both studied learning object-based, differentiable neural simulators. Their systems explicitly model the state of each object and learn to predict future states based on object interactions. In this paper, we combine such a learned object-based simulator with a physics engine for better prediction and for controlling real-world objects.

\subsection{Control with a Learned Simulator}

Recent papers have explored model-predictive control with deep networks~\cite{Lenz2015DeepMPC,Gu2016Continuous,nagabandi2017neural,farquhar2017treeqn,srinivas2018universal}. These approaches learn an abstract-state transition function, not an explicit model of the environment~\cite{Silver2017predictron,Oh2017Value}. Eventually, they apply the learned value function or model to guide policy network training. In contrast, we employ an object-based physical simulator that takes raw object states (\eg, velocity, position) as input. Hogan~\etal~\cite{hogan2018data} also learned a residual model with an analytical model for model-predictive control, but their learned model is a task-specific Gaussian Process, while our model has the ability to generalize to new object shapes and materials. 

A few papers have exploited the power of interaction networks for planning and control, mostly using interaction networks to help training policy networks via \emph{imagination}---rolling out approximate predictions~\cite{Racaniere2017Imagination,Hamrick2017Metacontrol,Pascanu2017Learning}. In contrast, we use interaction networks as a learned dynamics simulator, combine it with a physics engine, and directly search for actions in real-world control problems. Recently, Sanchez-Gonzalez~\etal~\cite{sanchez2018graph} also used interaction networks in control, though their model does not take into account explicit physical knowledge, and its performance is only demonstrated in simulation. 

\section{Method}
\label{sec:method}
In this section, we describe \model's formulation and components. We also present our Model Predictive Controller (MPC) which uses \model to perform the pushing task.

\begin{figure*}[t]
    \centering
    \includegraphics[width=.7\linewidth]{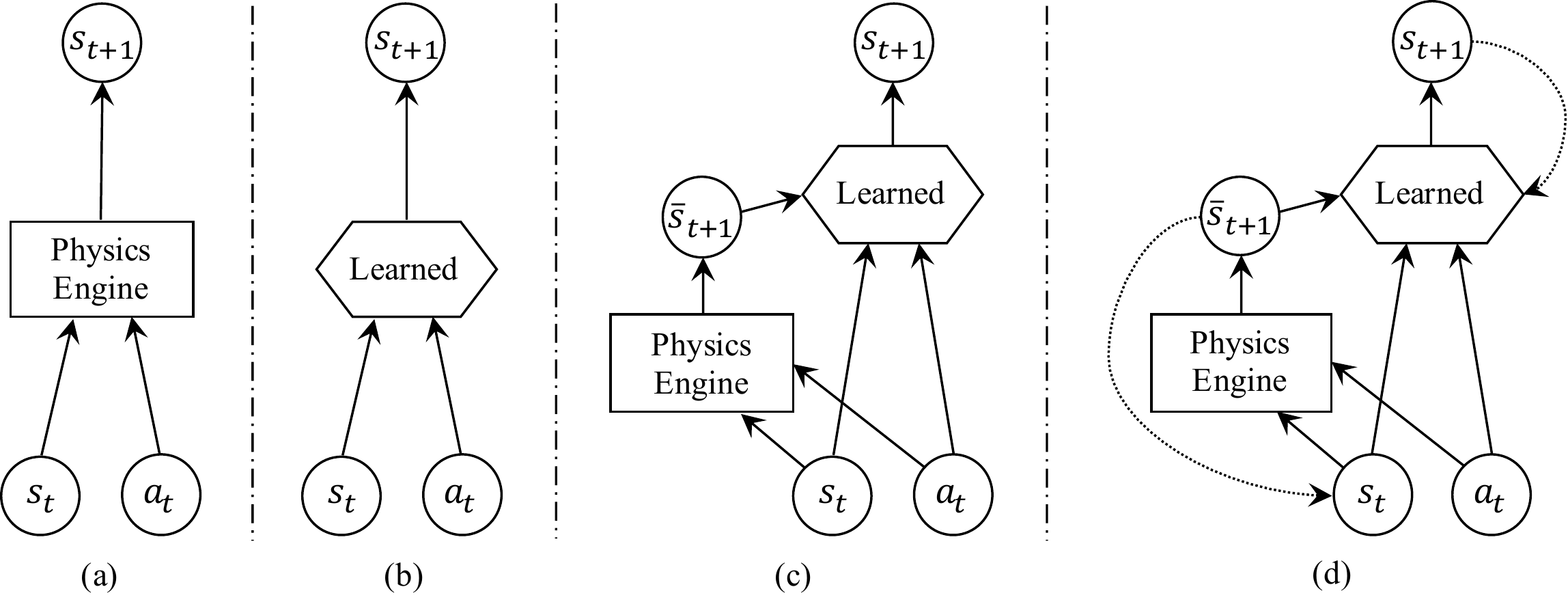}
    \caption{\textbf{Model classes:} (a) physics-based analytical models; (b) data-driven models; (c) simulator-augmented residual models; (d) recurrent simulator-augmented residual models.}
    \label{fig:formulation}
    \vspace{-15pt}
\end{figure*}

\subsection{Formulation}
Let $S$ be the state space and $A$ be the action space. A dynamics model is a function $f: S \times A \rightarrow S$ that predicts the next state given the current action and state: $f(s, a) \approx s', \;\;\text{for all}\;\; s, s' \in S, \;\; a \in A$. 

There are two general types of dynamics models: analytical (\fig{fig:formulation}a) and data-driven (\fig{fig:formulation}b). Our goal is to learn a hybrid dynamics model that combines the two (\fig{fig:formulation}c). Here, conditioned on the state-action pair, the data-driven model learns the discrepancy between analytical model predictions and real-world data (\ie the residual). Specifically, let $f_r$ represent the hybrid dynamics model, $f_p$ represent the physics engine, and $f_{\theta}$ represent the residual component. We have
%\begin{equation}
$f_r(s, a) = f_{\theta}(f_p(s,a), s, a) \approx s'$.
%\end{equation}
Intuitively, the residual model refines the physics engine's guess using the current state and action.%the recurrent parametric model. 
%\lpk{Whoa! What recurrent parametric model?  We need a description of the architecture, or a block diagram or something before saying this.}

%If $s_0$ is the initial state, $a_t$ the action at time $t$ and $\hat{s}_t$ the prediction at time $t$, then
%\begin{gather}
%f_{\theta}(\hat{s}_t, a_t) = \hat{s}_{t+1} \approx s_{t+1},\\
%f_{\theta}(s_0, a_0) = \hat{s}_1 \approx s_1,
%\end{gather}

For long-term prediction, let $f^R_{\theta}: S \times S \times A \rightarrow S$ represent the recurrent hybrid dynamics model (\fig{fig:formulation}d). If $s_0$ is the initial state, $a_t$ the action at time $t$, $\bar{s}_t$ the prediction by the physics engine $f_p$ at time $t$ and $\hat{s}_t$ the prediction at time $t$, then 
\begin{gather}
f^R_{\theta}(\bar{s}_{t+1}, \hat{s}_t, a_t) = \hat{s}_{t+1} \approx s_{t+1},\\
f_p(\bar{s}_t, a_t) = \bar{s}_{t+1}, \quad  \bar{s}_0 = \hat{s}_0 = s_0.
\end{gather}
For training, we collect observational data $\{(s_t, a_t, s_{t+1})\}_{t=0}^{T-1}$ and then solve the following optimization problem:
%\vspace{-5pt}
\begin{equation}
\theta^* = \argmin\limits_{\theta} \sum_{t=0}^{T-1} \|\hat{s}_{t+1} - s_{t+1}\|_2^2 + \lambda \|\theta\|_2^2,
%\vspace{-5pt}
\end{equation}
where $\lambda$ is the weight for the regularization term.

In this study, we choose to use a recurrent parametric model over a non-recurrent representation for two reasons. First, non-recurrent models are trained on observation data to make single-step predictions. Consequently, errors in prediction compound over a sequence of steps. Second, since these models recursively use their own predictions, the input data given during the simulation phase will have a different distribution than the input data during the training phase. This creates a data distribution mismatch between the training and test phases.
\vspace{-3pt}
\subsection{Interaction Networks}
We use {\em interaction networks}~\cite{Battaglia2016Interaction} as the data-driven model for multi-object interaction. An interaction network consists of 2 neural nets: $f_{\text{dyn}}$ and $f_{\text{rel}}$. The $f_{\text{rel}}$ network calculates pairwise forces between objects and the $f_{\text{dyn}}$ network calculates the next state of an object, based on the states of the objects it is interacting with and the nature of the interactions. 
%as a function of the pairwise interaction and the external force being applied on the current object. 
%\lpk{Not sure what "current object" means here.  Don't we want to say something like it predicts the next state of an object based on the states of the objects it is interacting with and the nature of the interactions?}

The original version of interaction networks was trained to make a single-step prediction; for improved accuracy, we extend them to make multi-step predictions. Let $s_t = \{o^1_t, o^2_t, \ldots, o^n_t\}$ be the state at time $t$, where $o^i_t$ is the state for object $i$ at time $t$. Similarly, let $\hat{s}_t = \{\hat{o}^1_t, \hat{o}^2_t, \ldots, \hat{o}^n_t\}$ be the predicted state at time $t$ where $o^i_t$ is the predicted state for object $i$ at time $t$. In our work, $o^i_t = [p^i_t, v^i_t, m^i, r^i]$ where $p^i_t$ is the pose of object $i$ at time step $t$, $v^i_t$ the velocity of object $i$ at time step $t$, $m^i$ the mass of object i and $r^i$ the radius of object i. Similarly, $\hat{o}^i_t = [\hat{p}^i_t, \hat{v}^i_t, m^i, r^i]$ where $\hat{p}^i_t$ is the predicted pose of object $i$ at time step $t$ and $\hat{v}^i_t$ the predicted velocity of object $i$ at time step $t$. Note that we do not predict any changes to static object properties such as mass and radius. Also, we note that while $s_t$ is a set of objects, the state of any individual object, $o^i_t$, is a vector. Now, let $a^i_t$ be the action applied to object $i$ at time $t$. The equations for the interaction network are:
\begin{align}
e^i_t & = \sum_{j \neq i} f_{\text{rel}}(v^i_t, p^i_t-p^j_t, v^i_t-v^j_t, m^i, m^j, r^i, r^j),\\
\hat{v}^i_{t+1} & = v^i_t + dt \cdot f_{\text{dyn}}(v^i_t, a^i_t, m^i, r^i, e^i_t),\\
\hat{p}^i_{t+1} & = p^i_t + dt \cdot \hat{v}^i_{t+1},\\
\hat{o}^i_{t+1} & = [\hat{p}^i_{t+1}, \hat{v}^i_{t+1}, m^i, r^i].
\end{align}
%\lpk{Better to use $\cdot$ than $\times$ unless you mean cartesian product or cross product or something.}
% lpk: repetitious
% These equations describe a single-step prediction. To do multi-step prediction, we can use the same equations by providing the true state $s_0$ at $t=0$ and predicted state $\hat{s}_t$ at $t > 0$ as inputs.

\subsection{\Modelfull (\model)}
A simulator-augmented interaction network extends an interaction network, where $f_{\text{dyn}}$ and $f_{\text{rel}}$ now take in the prediction of a physics engine, $f_p$.  We now learn the residual between the physics engine and the real world. Let $\bar{s}_t = \{\bar{o}^1_t, \bar{o}^2_t, \ldots, \bar{o}^n_t\}$ be the state at time $t$ and $\bar{o}^i_t$ be the state for object $i$ at time $t$ predicted by the physics engine. The equations for \model  are
\begin{align}
\bar{s}_{t+1} & = f_p(\bar{s}_t, a^1_t, a^2_t, \ldots, a^n_t),\\
e^i_t & = \sum_{j \neq i} f_{\text{rel}}(v^i_t, \bar{v}^i_{t+1}-\bar{v}^i_t, p^i_t-p^j_t, v^i_t-v^j_t, m^i, m^j, r^i, r^j),\\
\hat{v}^i_{t+1} & = v^i_t + dt \times f_{\text{dyn}}(v^i_t, \bar{p}^i_{t+1}-\bar{p}^i_t, a^i_t, m^i, r^i, e^i_t),\\
\hat{p}^i_{t+1} & = p^i_t + dt \times \hat{v}^i_{t+1},\\
\hat{o}^i_{t+1} & = [\hat{p}^i_{t+1}, \hat{v}^i_{t+1}, m^i, r^i].
\end{align}
These equations describe a single-step prediction. For multi-step prediction, we use the same equations by providing the true state $s_0$ at $t=0$ and predicted state $\hat{s}_t$ at $t > 0$ as input.

\subsection{Control Algorithm}
\label{sec:control_alg}
%We use $A^*$ search as our planning algorithm.   

Our action space has two free parameters: the point where the robot contacts the first disk and the direction of the push. In our experiments, a successful execution requires searching for a trajectory of about 50 actions. %For each time step, we discretize it the space into 6 possible contact points and 15 possible directions. \jw{Anurag, double check. Or maybe move this to experimental details.} %aour action space into we partition the range of each action parameter into a finite set of ranges and use the midpoints of those ranges as action choices.  %For the purposes of dynamic programming, we assume that two object states $o^i$ and $o^j$ are equal if the mass and radius in both the states are equal and if $\|p^i - p^j\| < \epsilon_p$ and $\|v^i - v^j\| < \epsilon_v$. We use the "Branch and Bound" technique to prune our search. 
Due to the size of the search space, we use an approximate receding horizon control algorithm with our dynamics model. The search algorithm maintains a priority queue of action sequences based on the heuristic below. For each expansion, let $s_t$ be the current state and $\hat{s}_{t+T}(a_t,\dots,a_{t+T-1})$ be the predicted state after $T$ steps with actions $a_t,\dots,a_{t+T-1}$. Let $s_g$ be the goal state. We choose the control strategy $a_t$ that minimizes the the cost function $||\hat{s}_{t+T}(s_t, a_t,\dots,a_{t+T-1}) - s_g||_2$ and insert the new action sequence into the queue.

\section{Experiments}
\label{sec:exp}

We demonstrate \model on a challenging planar manipulation task both in simulation and in the real-world. We further evaluate how our model generalizes to handle control tasks that involve objects of new materials and shapes.

\subsection{Task}

In this manipulation task, we are given two disks with different mass and radii. Our goal is to guide the second disk to a target location, but are constrained to push only the first disk. Here, a point $s$ in the state space is factored into a set of two object states, $s = \{o_1, o_2\}$, where each $o_i$ is an element of object state space $O$. The object state includes the mass, 2D position, rotation, velocity, and radius of the disk. 

Targets locations are generated at random and divide into two categories: easy and hard. A target location is produced by first sampling an angle $\alpha$ from an interval $U$, then choosing the goal location to be at distance of three times the radius of second disk and at an angle of $\alpha$ with respect to the second disk. In easy pushes, the interval $U$ is $[-\frac{\pi}{6}, \frac{\pi}{6}]$. In hard pushes, the interval $U$ is $[-\frac{\pi}{3}, -\frac{\pi}{6}] \cup [\frac{\pi}{6}, \frac{\pi}{3}]$. A push is considered a success if the distance between the goal location and the pose of the center of mass of the second disk is within $\frac{1}{10}^{th}$ the radius of second disk.

\subsection{Simulation Setup}

We use the Bullet physics engine \cite{Coumans2010Bullet} for simulation. 
%In simulation data, we fix the number of disks to 2. 
For each trajectory, we vary the coefficient of friction between the surface and the disks, the mass of the disks and their radius. The coefficient of friction is sampled from $\text{Uniform}(0.05, 0.25)$. The mass is sampled from Uniform(0.85kg, 1.15kg) and the radius is sampled from $\text{Uniform}(0.05\text{m}, 0.06\text{m})$. We always fix the initial position of the first disk to the origin. The other disks are placed in front of the first disk at an angle, randomly sampled from $\text{Uniform}(-\frac{\pi}{3}, \frac{\pi}{3})$, and just touches it. We ensure that disks don't overlap each other. The pusher is placed at back of the first disk at an angle randomly sampled from $\text{Uniform}(-\frac{\pi}{3}, \frac{\pi}{3})$, and just touches it. Then the pusher makes a straight line push at an angle, randomly sampled from $\text{Uniform}(-\frac{\pi}{6}, \frac{\pi}{6})$, for 2s and covers a distance of about 1cm. We experiment with two different simulation setups: (1) direct-force simulation setup in which we control pusher with external force and (2) robot control simulation setup in which we control the pusher using position control. We use the first setup to show the benefits of \model over other models. But in our real world setup, we control the pusher using position-based control. So, we have designed a second simulation setup which matches the real robot and use it to collect pre-training data. 

For the direct-force simulation setup, we collect $4500$ pushes with 2 disks  for our training set, $500$ pushes with 2 disks and $500$ pushes with 3 disks for our test set. For the robot control simulation setup, we collect $4500$ pushes with 2 disks for our training set and $500$ pushes with 2 disks for our test set. %The object state of disk consists of $2D$ position and angle about vertical axis, their respective velocities, mass and radius.

\subsection{Model and Implementation Details}
%Test set on simulation (Train points 4500 and test points 500)
% \begin{table*}[t]
%     \small
%     \centering
%     %\setlength{\tabcolsep}{2pt}
%     \begin{tabular}{lcccccc}
%     \toprule
%     \multirow{2}{*}{Models} & \multicolumn{3}{c}{Object 1} & \multicolumn{3}{c}{Object 2} \\
%     \cmidrule(lr){2-4}\cmidrule(lr){5-7} 
%     & trans (\%) &  pos (mm) & rot (deg) & trans (\%) & pos (mm) & rot (deg) \\
%     \midrule
%     Physics & 2.82 & 6.57 & 0.91 & 2.31 & 6.05 & 0.45 \\ 
%     IN & 2.09 & 5.61 & 0.68 & 1.47 & 3.79 & 0.26 \\
%     \model (ours) & {\bf 1.62} & {\bf 4.38} & {\bf 0.38} & {\bf 1.38} & {\bf 3.34} & {\bf 0.2} \\
%     \bottomrule
%     \end{tabular}
%     \caption{{\bf Errors on dynamics prediction in toy simulation setup}. \model achieves the best performance in both position and rotation estimation, compared with methods that rely on physics engines or neural nets alone. The first two metrics are the average Euclidean distance between the predicted and the ground truth object reported as a percentage relative to the initial pose (trans) and as absolute values (pos) in millimeters. The third is the average error of object rotation (rot) in degree. }
%     \label{tbl:toy_sim_pred}
% \end{table*}

\begin{table}[t]
    \small
    \centering
    \begin{tabular}{lccc}
    \toprule
    \multirow{2}{*}{Models} & \multicolumn{3}{c}{Error on Object 1/Object 2}\\
    \cmidrule(lr){2-4}
    & trans (\%) &  pos (mm) & rot (deg)\\
    \midrule
    Physics & 2.82/2.31 & 6.57/6.05 & 0.91/0.45\\ 
    IN & 2.09/1.47 & 5.61/3.79 & 0.68/0.26 \\
    \model (ours) & {\bf 1.62/1.38} & {\bf 4.38/3.34} & {\bf 0.38/0.2}\\
    \bottomrule
    \end{tabular}
    \caption{{\bf Errors on dynamics prediction in direct-force simulation setup}. \model achieves the best performance in both position and rotation estimation, compared with methods that rely on physics engines or neural nets alone. The first two metrics are the average Euclidean distance between the predicted and the ground truth object reported as a percentage relative to the initial pose (trans) and as absolute values (pos) in millimeters. The third is the average error of object rotation (rot) in degree. }
    \label{tbl:toy_sim_pred}
\end{table}

% \begin{table*}[t]
%     \small
%     \centering
%     \begin{tabular}{lcccccc}
%     \toprule
%     \multirow{2}{*}{Models} & \multicolumn{3}{c}{Object 1} & \multicolumn{3}{c}{Object 2} \\
%     \cmidrule(lr){2-4}\cmidrule(lr){5-7} 
%     & trans (\%) &  pos (mm) & rot (deg) & trans (\%) & pos (mm) & rot (deg) \\
%     \midrule
%     Physics & 2.82 & 6.57 & 0.91 & 2.31 & 6.05 & 0.45 \\ 
%     IN & 2.09 & 5.61 & 0.68 & 1.47 & 3.79 & 0.26 \\
%     \model (ours) & {\bf 1.62} & {\bf 4.38} & {\bf 0.38} & {\bf 1.38} & {\bf 3.34} & {\bf 0.2} \\
%     \bottomrule
%     \end{tabular}
%     \caption{{\bf Errors on dynamics prediction in toy simulation setup}. \model achieves the best performance in both position and rotation estimation, compared with methods that rely on physics engines or neural nets alone. The first two metrics are the average Euclidean distance between the predicted and the ground truth object reported as a percentage relative to the initial pose (trans) and as absolute values (pos) in millimeters. The third is the average error of object rotation (rot) in degree. }
%     \label{tbl:toy_sim_pred}
% \end{table*}
We compare two models for simulation and control: the original interaction networks (IN) and our \modelfull (\model). They share the same architecture. Each consists of two separate neural networks: $f_{\text{dyn}}$ and $f_{\text{rel}}$. Both $f_{\text{dyn}}$ and $f_{\text{rel}}$ have four linear layers with hidden sizes of 128, 64, 32 and 16 respectively. The first three linear layers are followed by a ReLU activation.

%\subsection{Training Details}
%\input{figText/table_sim_control.tex}

%We first collect data using a random policy in disk pushing task in simulation. Then we use the simulation data to train neural nets, serving as our learned dynamics models. We use our planner with each of these models and compare their performance in disk pushing task in simulation. After that, we use the model, with best performance in simulation, with our planner to perform disk pushing task in real world and collect the real world data in the process. We fine-tune the model on real world data and again use them with our planner to perform disk pushing task in real world. We compare performance of the model before and after finetuning.

%For simulated data, we fix number of disks to 2 and we vary their mass and radius. We also vary the coefficient of friction between the disks and the surface but the Interaction network isn't conditioned on this coefficient of friction. So, we are implicitly training an ensemble of Interaction networks and our final model is mean of these ensemble models. There are two reasons behind varying the coefficient of friction but not conditioning our models on them. First, the coefficient of friction often varies across space on the same surface and therefore, it's hard to calculate the true coefficient of friction. Second, this kind of training leads to a better transfer from simulation to real world. 

Training interaction networks in simulation is straightforward. It is more involved for \model, which learns a correction over the Bullet physics engine, so the problem of training ``in simulation'' is ill-posed. To address this problem, we fix the physics engine in \model with mass and radius of disks equaling that of disks in the real world. We also fix the coefficient of friction in the physics engine to an estimated mean of the coefficient of friction of the real world surface across space. The training data instead contain varied mass and radius of both disks, and varied the coefficient of friction between the disks and the surface, and the model is trained to learn the residual. %By varying mass, radius and the coefficient of friction in the second physics engine, we hope that the distribution of residuals generated either contains or is close to the residual between the physics engine and the real world.   

%Note that the residual Interaction network component is still not conditioned on coefficient of friction and therefore is implicitly a mean of ensemble models. 
%After we obtain the models trained in simulation, we use them separately to perform our task and record the real-world data. Finally, we use this real world data to fine-tune these models. 

We use ADAM~\cite{Kingma2015Adam} for optimization with a starting learning rate of 0.001. We decrease it by 0.5 every 2,500 iterations. We train these models for 10,000 iterations with a batch size of 100. Let the predicted 2D position, rotation, and velocity at time $t$ of disk $i$ be $\hat{p}^i_t$, $\hat{r}^i_t$ and $\hat{v}^i_t$, respectively, and the corresponding true values be $p^i_t$, $r^i_t$, and $v^i_t$. Let $T$ be the length of all trajectories. The training loss function for a single trajectory is 
\begin{dmath}
    \frac{1}{T}\sum_{i=1}^2\sum_{t=0}^{T-1} \|p^i_t - \hat{p}^i_t\|_2^2 + \|v^i_t - \hat{v}^i_t\|_2^2 + \\
    \|\sin{(r^i_t)} - \sin{(\hat{r}^i_t)}\|_2^2 + \|\cos{(r^i_t)} - \cos{(\hat{r}^i_t)}\|_2^2.
\end{dmath}
During training, we use a batch of trajectories and take a mean over the loss of those trajectories. We also use $l_2$ regularization with $10^{-3}$ as regularization constant.

In practice, we ask the models to predict the change in object states (relative values) rather than the absolute values. This enables them to generalize to arbitrary starting positions without overfitting.

\subsection{Search Algorithm}
% 500 test points  
% \begin{table*}[t]
%     \small
%     \centering
%     %\setlength{\tabcolsep}{2pt}
%     \begin{tabular}{lccccccccc}
%     \toprule
%     \multirow{2}{*}{Models} & \multicolumn{3}{c}{Object 1} & \multicolumn{3}{c}{Object 2} & \multicolumn{3}{c}{Object 3} \\
%     \cmidrule(lr){2-4}\cmidrule(lr){5-7} 
%     & trans (\%) &  pos (mm) & rot (deg) & trans (\%) & pos (mm) & rot (deg) & trans (\%) & pos (mm) & rot (deg)\\
%     \midrule
%     Physics & 2.79 & 6.53 & 0.89 & 2.34 & 6.11 & 0.48 & 2.38 & 6.21 & 0.49 \\ 
%     IN & 2.12 & 5.68 & 0.70 & 1.63 & 4.41 & 0.34 & 1.67 & 4.52 & 0.38\\
%     \model (ours) & {\bf 1.68} & {\bf 4.54} & {\bf 0.41} & {\bf 1.52} & {\bf 3.97} & {\bf 0.25} & {\bf 1.61} & {\bf 4.34} & {\bf 0.32} \\
%     \bottomrule
%     \end{tabular}
%     \caption{{\bf Generalization to 3 objects in toy simulation setup}. \model achieves the best generalization in both position and rotation estimation}
%     \label{tbl:toy_sim_gen}
% \end{table*}

\begin{table}[t]
    \small
    \centering
    \setlength{\tabcolsep}{5pt}
    \begin{tabular}{lccc}
    \toprule
    \multirow{2}{*}{Models} & \multicolumn{3}{c}{Error on Object 1/2/3}\\
    \cmidrule(lr){2-4}
    & trans (\%) &  pos (mm) & rot (deg)\\
    \midrule
    Physics & 2.79/2.34/2.38 & 6.53/6.11/6.21 & 0.89/0.48/0.49 \\ 
    IN & 2.12/1.63/1.67 & 5.68/4.41/4.52 & 0.70/0.34/0.38 \\
    \model (ours) & {\bf 1.68/1.52/1.61} & {\bf 4.54/3.97/4.34} & {\bf 0.41/0.25/0.32}\\
    \bottomrule
    \end{tabular}
    \caption{{\bf Generalization to 3 objects in direct-force simulation setup}. \model achieves the best generalization in both position and rotation estimation.}
    \label{tbl:toy_sim_gen}
    \vspace{-5pt}
\end{table}

\begin{table}[t]
    \small
    \centering
    \begin{tabular}{lcccccc}
    \toprule
    \multirow{2}{*}{Models} & \multicolumn{3}{c}{Error on Object 1/2}\\ 
    \cmidrule(lr){2-4}
    & trans (\%) &  pos (mm) & rot (deg)\\
    \midrule
    Physics & 2.52/2.19 & 6.27/5.81 & 0.85/0.29\\ 
    IN & 2.13/1.59 & 5.76/3.84 & 0.72/0.28\\
    \model (ours) & {\bf 1.82/1.50} & {\bf 4.66/3.47} & {\bf 0.40/0.21}\\
    \bottomrule
    \end{tabular}
    \caption{{\bf Errors on dynamics prediction in robot control simulation setup}. \model achieves the best performance in both position and rotation estimation.}
    \label{tbl:sim_pred}
    \vspace{-15pt}
\end{table}

% \begin{table*}[t]
%     \small
%     \centering
%     %\setlength{\tabcolsep}{2pt}
%     \begin{tabular}{lcccccc}
%     \toprule
%     \multirow{2}{*}{Models} & \multicolumn{3}{c}{Object 1} & \multicolumn{3}{c}{Object 2} \\
%     \cmidrule(lr){2-4}\cmidrule(lr){5-7} 
%     & trans (\%) &  pos (mm) & rot (deg) & trans (\%) & pos (mm) & rot (deg) \\
%     \midrule
%     Physics & 2.52 & 6.27 & 0.85 & 2.19 & 5.81 & 0.29 \\ 
%     IN & 2.13 & 5.76 & 0.72 & 1.59 & 3.84 & 0.28 \\
%     \model (ours) & {\bf 1.82} & {\bf 4.66} & {\bf 0.40} & {\bf 1.50} & {\bf 3.47} & {\bf 0.21} \\
%     \bottomrule
%     \end{tabular}
%     \caption{{\bf Errors on dynamics prediction in pre-train simulation setup}. \model achieves the best performance in both position and rotation estimation}
%     \label{tbl:sim_pred}
% \end{table*}

%We use $A^*$ search with "Branch and Bound" as our planner. We set $\epsilon_v$ and $\epsilon_p$ to $1E-3$. 
As mentioned in \sect{sec:control_alg}, an action is defined by the initial position of the pusher and the angle of the push, $\alpha$, with respect to the first disk. After these parameters have been selected, the pusher starts at the initial position and moves at an angle of $\alpha$ with respect to the first disk for $10mm$. We discretize our action space as follows. For selecting $\alpha$, we divide the interval $[\frac{-\pi}{6}, \frac{\pi}{6}]$ into six bins and choose their midpoints. For selecting the initial position of the pusher, we choose an angle $\theta$ and place the pusher at edge of first disk at an angle $\theta$ such that the pusher touches the first disk. For selecting $\theta$, we divide the interval $[\frac{-\pi}{3}, \frac{\pi}{3}]$ into 12 bins and choose midpoint of one of these bins. Therefore, our action space consists of 72 discretized actions for each time step. We maintain a priority queue of action sequences based on heuristic $h(\hat{p}^1, \hat{p}^2, p_g)$ where $\hat{p}^i$ is the predicted 2D position of disk $i$ and $p_g$ is the 2D position of goal. $h(\hat{p}^1, \hat{p}^2, p_g)$ is sum of $\|\hat{p}^2 - p_g\|$ and cosine distance between $p_g - \hat{p}^1$ and $\hat{p}^2 - \hat{p}^1$. The cosine distance serves as a regularization cost to encourage the center of both disks and the goal to stay in a straight line. To prevent the priority queue from blowing up, we do receding horizon greedy search with an initial horizon of 2, and increase it to 3 when the distance between the second disk and goal is less than $10mm$.
%Test set on real data (Train points 1200 and test points 300)

%* means no finetuning
% \begin{table*}[t]
%     \small
%     \centering
%     \begin{tabular}{lccccccc}
%     \toprule
%     \multirow{2}{*}{Models} & \multirow{2}{*}{Fine-tuning} & \multicolumn{3}{c}{Object 1} & \multicolumn{3}{c}{Object 2} \\
%     \cmidrule(lr){3-5}\cmidrule(lr){6-8} 
%     & & trans (\%) &  pos (mm) & rot (deg) & trans (\%) & pos (mm) & rot (deg) \\
%     \midrule
%     Physics & N/A & 0.87 & 3.06 & {\bf 0.32} & 1.91 & 6.41 & {\bf 0.17} \\ 
%     IN & No & 0.86 & 2.96 & 0.96 & 1.84 & 5.75 & 0.32 \\
%     \model (ours) & No & 0.69 & 2.38 & 0.43 & 1.06 & 3.52 & 0.18 \\
%     \midrule
%     IN & Yes & 0.63 & 2.23 & 0.41 & 0.61 & 2.05 & 0.19 \\
%     \model (ours) & Yes & {\bf 0.42} & {\bf 1.50} & 0.34 & {\bf 0.43} & {\bf 1.52} & {\bf 0.17} \\
%     \bottomrule
%     \end{tabular}
%     \caption{{\bf Errors on dynamics prediction in the real world.} \model obtains the best performance in both position and rotation estimation. Its performance gets further improved after fine-tuning on real data.}
%     \label{tbl:real_pred}
% \end{table*}

\begin{table}[t]
    \small
    \centering
    \setlength{\tabcolsep}{5pt}
    \begin{tabular}{lcccc}
    \toprule
    \multirow{2}{*}{Models} & \multirow{2}{*}{Fine-tuning} & \multicolumn{3}{c}{Error on Object 1/2}\\
    \cmidrule(lr){3-5}
    & & trans (\%) &  pos (mm) & rot (deg)\\
    \midrule
    Physics & N/A & 0.87/1.91 & 3.06/6.41 & {\bf 0.32/0.17}\\ 
    IN & No & 0.86/1.84 & 2.96/5.75 & 0.96/0.32 \\
    \model (ours) & No & 0.69/1.06 & 2.38/3.52 & 0.43/0.18\\
    \midrule
    IN & Yes & 0.63/0.61 & 2.23/2.05 & 0.41/0.19 \\
    \model (ours) & Yes & {\bf 0.42/0.43} & {\bf 1.50/1.52} & 0.34/{\bf   0.17}\\
    \bottomrule
    \end{tabular}
    \caption{{\bf Errors on dynamics prediction in the real world.} \model obtains the best performance in both position and rotation estimation. Its performance gets further improved after fine-tuning on real data.}
    \label{tbl:real_pred}
    \vspace{-15pt}
\end{table}

\subsection{Prediction and Control Results in Simulation}
\begin{figure}[t]
    \centering
    \begin{subfigure}[b]{\linewidth}
        \includegraphics[width=.32\linewidth]{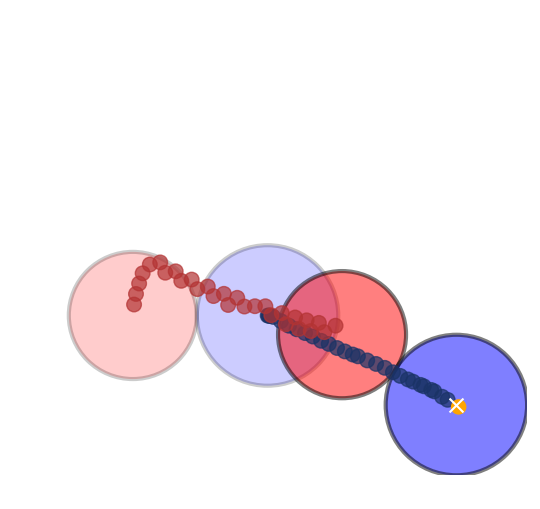}
        \includegraphics[width=.32\linewidth]{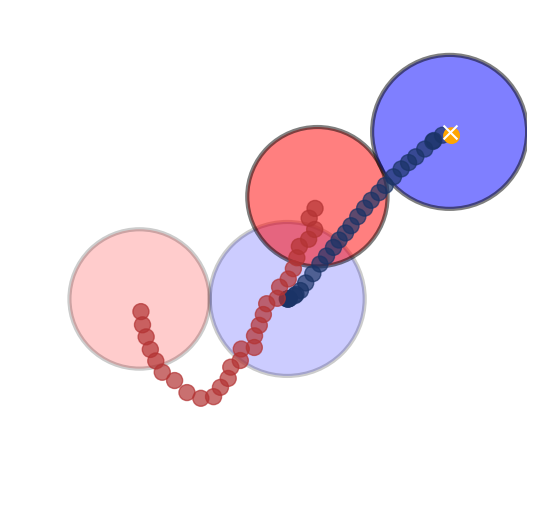}
        \includegraphics[width=.32\linewidth]{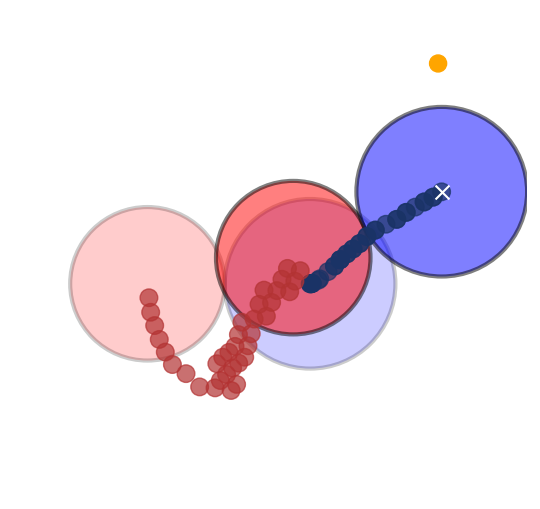}
        \caption{\label{fig:neural} Control with interaction networks (IN)}
    \end{subfigure}
    \begin{subfigure}[b]{\linewidth}
        \includegraphics[width=.32\linewidth]{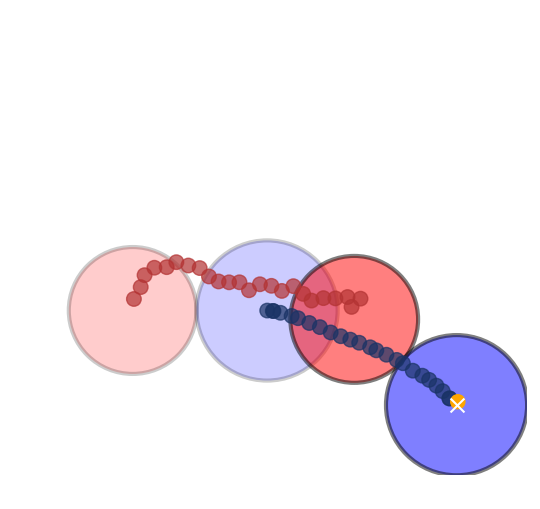}
        \includegraphics[width=.32\linewidth]{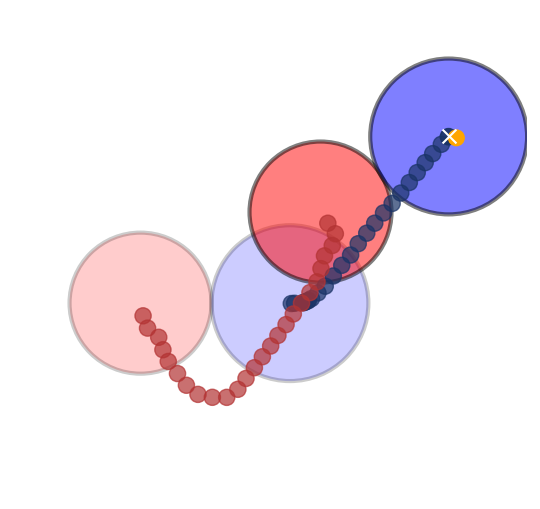}
        \includegraphics[width=.32\linewidth]{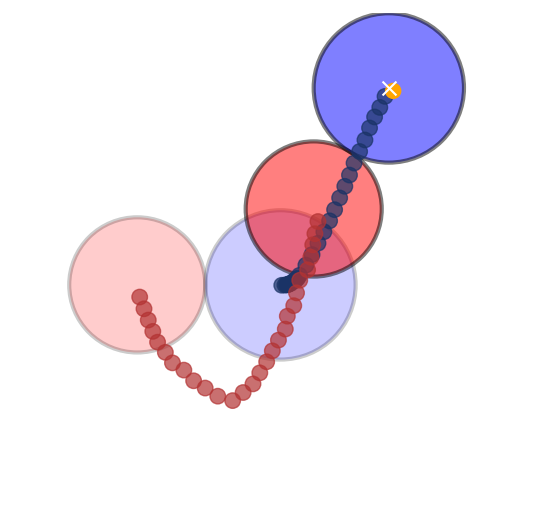}
        \caption{\label{fig:hybrid} Control with \modelfull(\model)}
    \end{subfigure}
    \ \\
    \vspace{3pt}
    
    \small
    \begin{tabular}{lcc}
    \toprule
    Models & Easy Push & Hard Push \\
    \midrule
    IN & 100\% & 88\% \\
    \model (ours) & 100\% & 100\% \\
    \bottomrule
    \end{tabular}
    \caption{{\bf Qualitative results and success rates on control tasks in simulation.} The goal is to push the red disk so that the center of the blue disk reaches the target region (yellow). The transparent shapes are their initial positions, and the solid ones are their final positions after execution. The center of the blue disk after execution is marked as a white cross. (a) Control with interaction networks works well but makes mistakes occasionally (column 3). (b) Control with \model achieves perfect performance in simulation.}
    \label{fig:sim_control}
    \vspace{-15pt}
\end{figure}

The forward multi-step prediction errors of both interaction networks and \model for direct-force simulation setup with 2 disks and 3 disks are reported in \tbl{tbl:toy_sim_pred} and \tbl{tbl:toy_sim_gen}. Note that errors on different objects are separated by $/$ in all the tables. The training data for this setup consist of pushes with only 2 disks. The forward multi-step prediction errors of both interaction networks and \model for robot control simulation setup are reported in \tbl{tbl:sim_pred}. Given an initial state and a sequence of actions, the models do forward prediction for the next 200 timesteps, where each time-step is 1/240s. We see that \model outperforms interaction networks in both setups. We also list the results of the fixed physics engine used for training \model for reference. 

% The forward multi-step prediction errors of both interaction networks and \model in simulation are reported in \tbl{tbl:sim_pred}. Given an initial state and a sequence of actions, the models do forward prediction for next 200 timesteps, where each time-step is 1/240s. We see that \model outperforms interaction networks. We also list the results of the fixed physics engine used for training \model for reference. %has the highest prediction error, it is not a fair comparison because physics engine (of augmented Interaction networks) has the wrong parameters. The mass and radius of disks in the physics engine is same as that in the real world. But, in simulated data, we randomly sample mass and radius of the disks. 

We have also evaluated IN and \model on control tasks in simulation. We test each model on 25 easy and 25 hard pushes. For these pushes, we set the mass of two disks to 0.9kg and 1kg and their radius to 54mm and 59mm, making them differ them from those used in the internal physics engine of \model. This mimics real-world environments, where objects' size and shape cannot be precisely estimated, and ensures \model cannot cheat by just querying the internal simulator. \fig{fig:sim_control} shows \model performs better than IN. This suggests learning the residual not only helps to do better forward prediction, but also benefits control.

%As the disks in \model's physics engine of \model has same mass and radius as the real world. Therefore, to test the /model in simulation for control tasks, we choose mass of two disks as 0.9kg and 1kg and their radius as 54mm and 59mm. This is to ensure that the physics engine of \model isn't same as the true model of simulation. 

\vspace{-1pt}
\subsection{Real-World Robot Setup}
\vspace{-1pt}

\begin{figure}[t]
    \centering
    \begin{subfigure}[b]{\linewidth}
        \includegraphics[width=.32\linewidth]{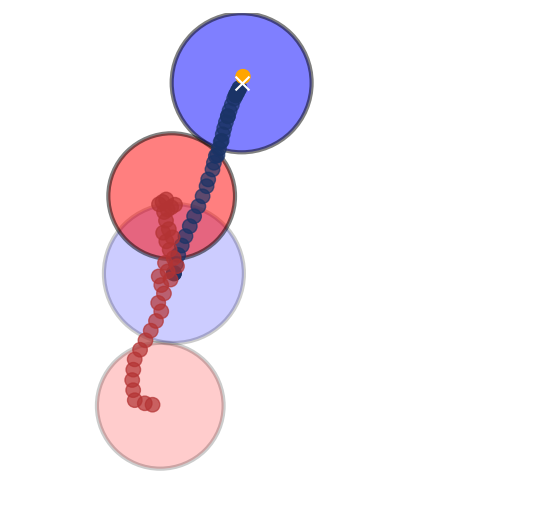}
        \includegraphics[width=.32\linewidth]{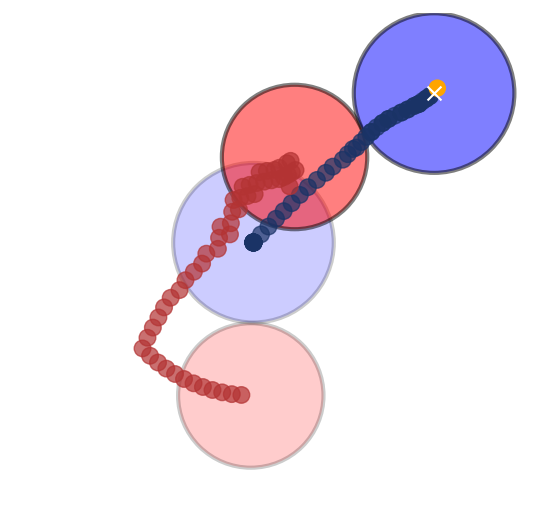}
        \includegraphics[width=.32\linewidth]{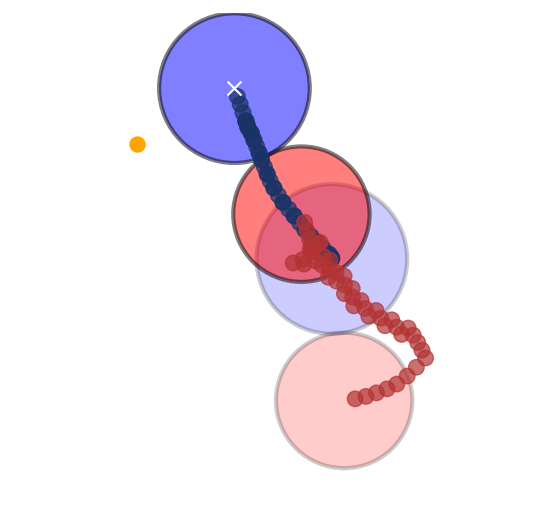}
        \caption{\label{fig:sim_only} Using the model trained on simulated data only}
    \end{subfigure}
    \hfill
    \begin{subfigure}[b]{\linewidth}
        \includegraphics[width=.32\linewidth]{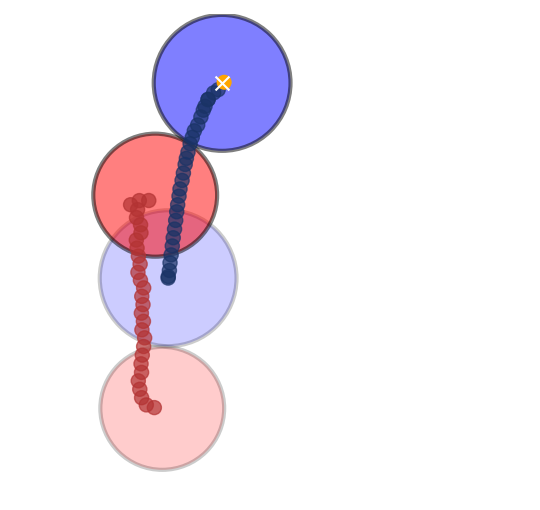}
        \includegraphics[width=.32\linewidth]{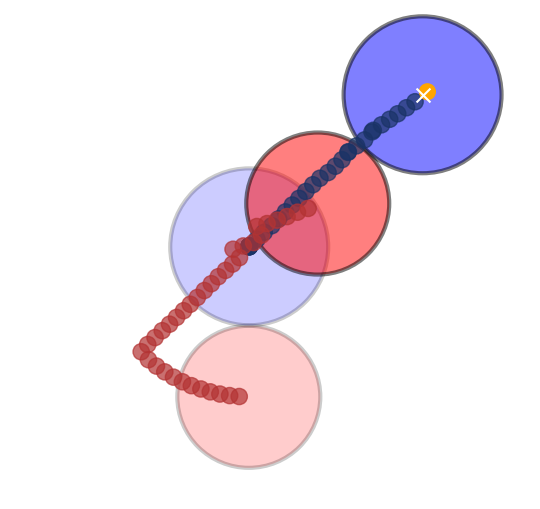}
        \includegraphics[width=.32\linewidth]{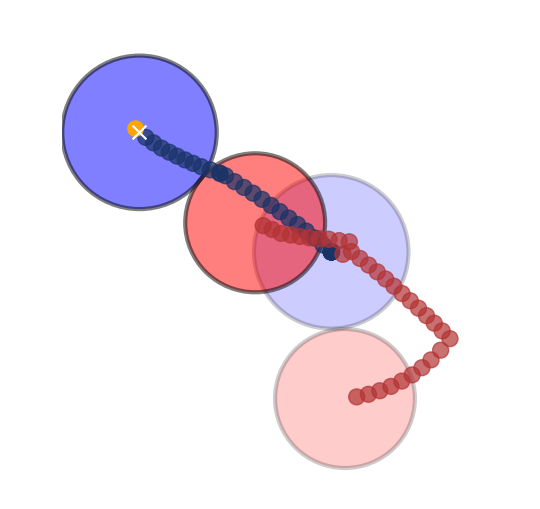}
        \caption{\label{fig:real_sim} Using the model trained on both simulated and real data}
    \end{subfigure}
    \small
    \begin{tabular}{llcc}
    \toprule
    Model & Data & Easy Push & Hard Push \\
    \midrule
    \model & Sim & 100\% & 68\%\\
    \model & Sim + Real & 100\% & 96\% \\
    \bottomrule
    \end{tabular}
    \caption{{\bf Qualitative results and success rates on control tasks in the real world.} The goal is to push the red disk so that the center of the blue disk reaches the target region (yellow). The left column is an example of easy pushe, while the right two show hard pushes. (a) The model trained on simulated data only performs well for easy pushes, but sometimes fails on harder control tasks; (b) The model trained on simulated and real data improves the performance, working well for both easy and hard pushes.}
    \label{fig:real_control}
    \vspace{-15pt}
\end{figure}
\begin{figure}[t]
    \centering
    \begin{subfigure}[b]{\linewidth}
        \includegraphics[width=.32\linewidth]{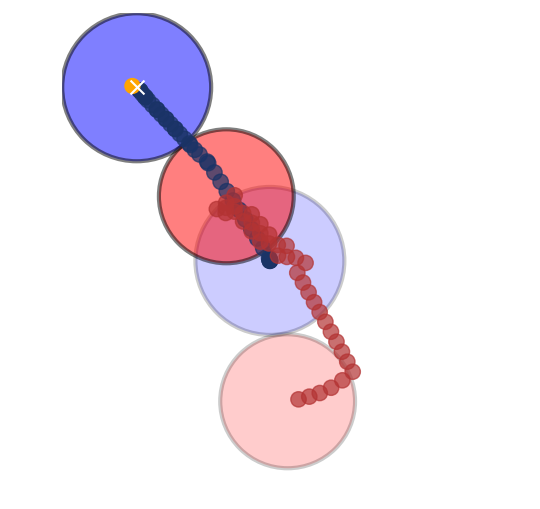}        
        \includegraphics[width=.32\linewidth]{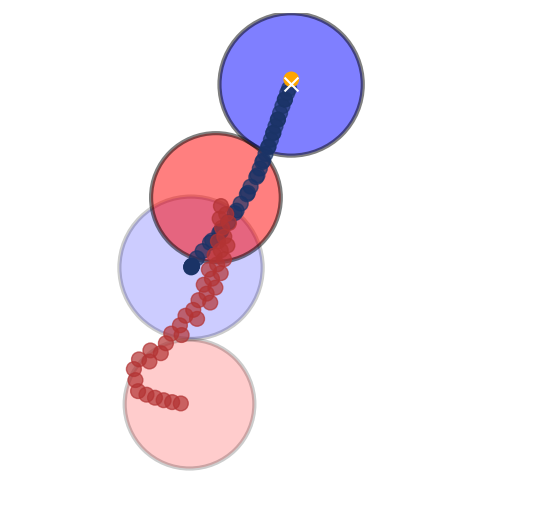}        
        \includegraphics[width=.32\linewidth]{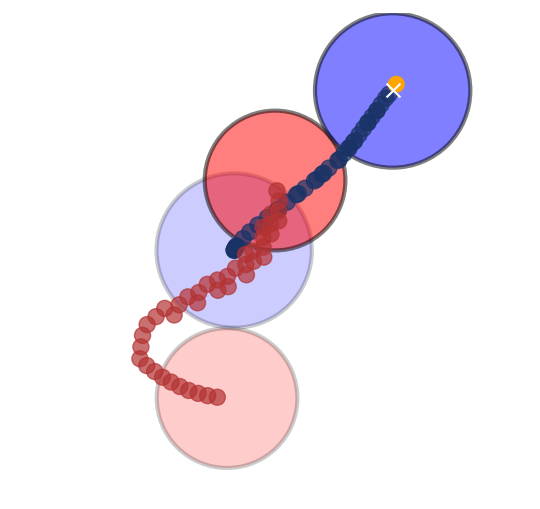}        
        \caption{\label{fig:mat_gen} Results on a new surface}
    \end{subfigure}
    \hfill
    \begin{subfigure}[b]{\linewidth}
        \includegraphics[width=.32\linewidth]{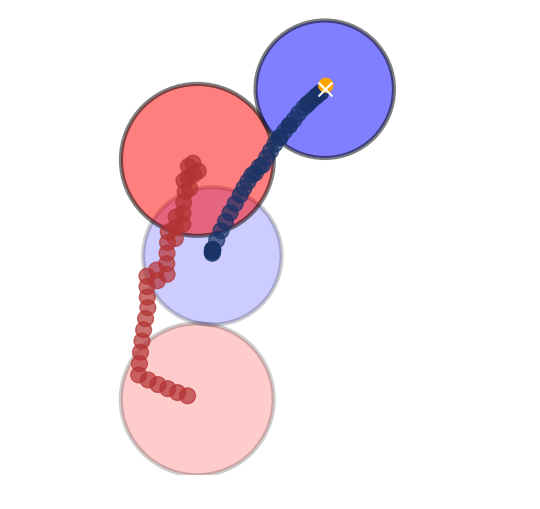}
        \includegraphics[width=.32\linewidth]{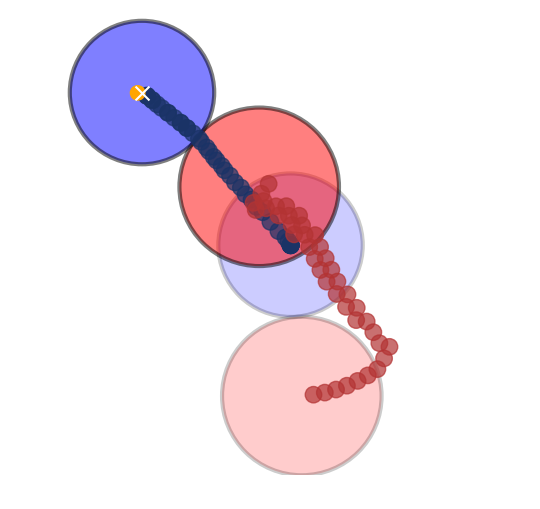}
        \includegraphics[width=.32\linewidth]{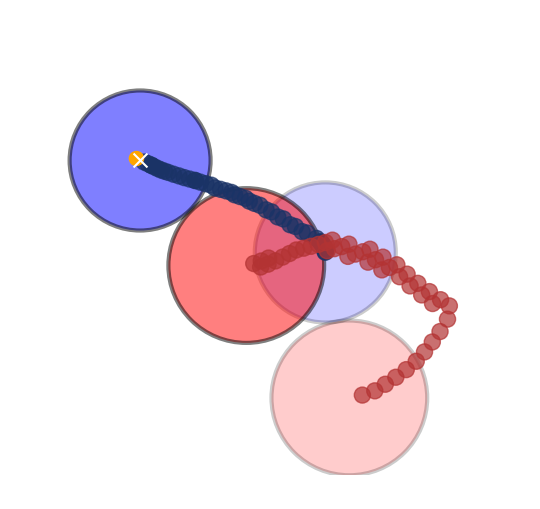}
        \caption{\label{fig:shape_gen} Results on a new shape with a smaller radius}
    \end{subfigure}
    \caption{{\bf Generalization to new scenarios}. (a) \model trained generalizes to control tasks on a new surface (plywood), with a success rate of 92\%; (b): It also generalizes to tasks where the two disks have a different radius (red: small $\rightarrow$ large; blue: large $\rightarrow$ small), with a success rate of 88\%. All tasks here are hard pushes.}
    \label{fig:generalize}
    \vspace{-15pt}
\end{figure}

We now test our models on a real robot. The setup used for the real experiments is based on the system from the MIT Push dataset~\cite{Yu2016More}. The pusher is a cylinder of radius 4.8mm attached to the last joint of a ABB IRB 120 robot. The position of the pusher is recorded using the robot kinematics. The two disks being pushed are made of stainless steel, have radius of 52.5mm and 58mm, and weight 0.896kg and 1.1kg. During the experiments, the smallest disk is the one pushed directly by the pusher. The position of both disks is tracked using a Vicon system of four cameras so that the disks' positions are highly accurate. Finally, the surface where the objects lay is made of ABS (Acrylonitrile Butadiene Styrene), whose coefficient of friction is around 0.15. Each push is done at 50mm/s and spans 10mm. We collect 1,500 pushes out of which 1,200 are used for training and 300 for testing.

We evaluate two versions of interaction networks and \model. The first is an off-the-shelf version purely trained on synthetic data; the second is one trained on simulated data and later fine-tuned on real data. This helps us understand whether these models can exploit real data to adapt to new environments. % Finally, we test the fine-tuned model on easy and hard pushes.
\vspace{-1pt}
\subsection{Results on Real-World Data}
\vspace{-1pt}
Results of forward simulation are shown in \tbl{tbl:real_pred}. \model outperforms IN on real data. While both models benefit from fine-tuning, \model achieves the best performance. This suggests residual learning also generalizes to real data well. All models achieve a lower error on real data than in simulation; this is because simulated data have a significant amount of noise to make the problem more challenging.

%The prediction error of physics engine and augmented Interaction networks before and after fine-tuning with real world data is shown in \tbl{}. As expected, the augmented Interaction networks after finetuning has the least prediction error.

We then evaluate \model (both with and without fine-tuning) for control, on 25 easy and 25 hard pushes. The results are shown in \fig{fig:real_control}. The model without fine-tuning achieves 100\% success rate on easy pushes and 68\% on hard pushes. As shown in the rightmost columns of \fig{fig:sim_only}, it sometimes pushes the object too far and gets stuck in a local minimum. After fine-tuning, the model works well on both easy pushes (100\%) and hard pushes (96\%) (\fig{fig:real_sim}).

While objects of different shapes and materials have different dynamics, the gap between their dynamics in simulation and in the real world might share similar patterns. This is the intuition behind the observation that residual learning allows easier generalization to novel scenarios. Ajay~\cite{ajay2018augmenting} validated this for forward prediction. Here, we evaluate how our fine-tuned \model generalizes for control. We test our model on 25 hard pushes with a different surface (plywood, where the original surface is ABS), using the original disks. Our framework achieves successes in 92\% of the pushes, where \fig{fig:mat_gen} shows qualitative results.  %Our \model learns a correction over a simulator. Thus, even if we change the parameters of our simulator (like radius, mass of disks) but the residual between simulator and real world remain same, our model should still be able to work. To test this, we changed radius of disks to 58mm and 52.5mm. The order of the disks matter because we push the first disk. Then, 
We've also evaluated our model on another 25 hard pushes, where it pushes the large disk (58mm) to direct the small one (52.5mm). Our framework achieves successes in 88\% of the pushes. \fig{fig:shape_gen} shows qualitative results. These results suggest that \model can generalize to solve control tasks with new object shapes and materials.

\vspace{-1pt}
\section{Conclusion}
\vspace{-1pt}
\label{sec:discussion}

We have proposed a hybrid dynamics model, \modelfull (\model), combining a physical simulator with a learned, object-centered neural network. Our underlying philosophy is to first use analytical models to model real-world processes as much as possible, and learn the remaining residuals. Learned residual models are specific to the real-world scenario for which data is collected, and adapt the model accordingly. The combined physics engine and residual model requires little need for domain specific knowledge or hand-crafting and can generalize well to unseen situations. We have demonstrated \model's efficacy when applied to a challenging control problem in both simulation and the real world. Our model also generalizes to setups where object shape and material vary and has potential applications in control tasks that involve complex contact dynamics.

\noindent {\bf Acknoledgements. } This work is supported by NSF \#1420316, \#1523767, and \#1723381, AFOSR grant FA9550-17-1-0165, ONR MURI N00014-16-1-2007, Honda Research, Facebook, and Draper Laboratory. 

\bibliographystyle{IEEEtran}
\bibliography{physplus}

\end{document}